\pdfoutput=1 
\documentclass[10pt,twocolumn,letterpaper]{article}

\usepackage{iccv}
\usepackage{times}
\usepackage{epsfig}
\usepackage{graphicx}
\usepackage{amsmath}
\usepackage{amssymb}

\usepackage{booktabs}
\usepackage{mathtools,graphicx}
\usepackage{multirow}
\newcommand\vertarrowbox[3][6ex]{%
  \begin{array}[t]{@{}c@{}} #2 \\
  \left\uparrow\vcenter{\hrule height #1}\right.\kern-\nulldelimiterspace\\
  \makebox[0pt]{\scriptsize#3}
  \end{array}%
}

\usepackage[pagebackref=true,breaklinks=true,letterpaper=true,colorlinks,bookmarks=false]{hyperref}

\iccvfinalcopy 

\def\iccvPaperID{1861} 

\ificcvfinal\fi
\usepackage{enumitem,bigdelim}
\usepackage{adjustbox}
\usepackage[dvipsnames]{xcolor}
\usepackage{xspace}

\newcommand{\MakeArrowVar}[1]{\multicolumn{#1}{c}{\upbracefill \hspace*{0.0em}}}

\newcommand{\boldparagraph}[1]{\vspace{0.2em}\noindent{\bf #1} }    

\definecolor{custom_red}{rgb}{0.8500,0.3250,0.0980}%
\definecolor{custom_green}{rgb}{0.4660,0.6740,0.1880}%
\definecolor{custom_blue}{rgb}{0,0.4470,0.7410}%

\newcommand{\im}{{I}}
\newcommand{\bgr}{{B}}

\newcommand{\tm}{i}
\newcommand{\tmbr}{_\tm}
\newcommand{\rot}{R\tmbr}
\newcommand{\tran}{T\tmbr}
\newcommand{\pose}{\theta\tmbr}
\newcommand{\shape}{\beta}
\newcommand{\texture}{\mathcal{T}}

\newcommand{\motion}{\mathcal{M}}
\newcommand{\meshi}[1]{\Theta_{#1}}
\newcommand{\mesh}{\meshi{\tmbr}}

\newcommand{\renf}{\mathcal{R}_F}
\newcommand{\rens}{\mathcal{R}_S}

\newcommand{\alpmat}{\alpha_{\text{in}}}

\newcommand{\coeff}{\boldsymbol{C}}
\newcommand{\indic}{\boldsymbol{I_c}}

\def\ie{\emph{i.e}\onedot} 
\usepackage{lipsum}

\usepackage[capitalize]{cleveref}
\crefname{section}{Sec.}{Secs.}
\Crefname{section}{Section}{Sections}
\Crefname{table}{Table}{Tables}
\crefname{table}{Tab.}{Tabs.}

\usepackage[accsupp]{axessibility}  

\begin{document}

\title{Human from Blur: Human Pose Tracking from Blurry Images}
\ificcvfinal\thispagestyle{empty}\fi

\author{
Yiming Zhao$^{1}$ \qquad Denys Rozumnyi$^{1}$\qquad Jie Song$^{1}$  \\
Otmar Hilliges$^{1}$  \qquad Marc Pollefeys$^{1,2}$  \qquad Martin R. Oswald$^{1,3}$ \\ \vspace{0.5em}
$^{1}$ETH Z\"urich\qquad
$^{2}$Microsoft\qquad $^{3}$University of Amsterdam
}

\maketitle

\begin{abstract}

We propose a method to estimate 3D human poses from substantially blurred images.
The key idea is to tackle the inverse problem of image deblurring by modeling the forward problem with a 3D human model, a texture map, and a sequence of poses to describe human motion.
The blurring process is then modeled by a temporal image aggregation step.
Using a differentiable renderer, we can solve the inverse problem by backpropagating the pixel-wise reprojection error to recover the best human motion representation that explains a single or multiple input images.
Since the image reconstruction loss alone is insufficient, we present additional regularization terms.
To the best of our knowledge, we present the first method to tackle this problem.
Our method consistently outperforms other methods on significantly blurry inputs since they lack one or multiple key functionalities that our method unifies, i.e.~image deblurring with sub-frame accuracy and explicit 3D modeling of non-rigid human motion. 
\end{abstract}


\section{Introduction}  \label{sec:intro}

\thispagestyle{empty}

Accurate tracking of human motion is often crucial for understanding dynamic scenes from images.
Human motion estimation has a wide field of applications such as improving human-robot collaboration~\cite{al2020real}, human-machine interaction in general~\cite{heindl2019metric}, better safety for autonomous driving~\cite{kress2018human}, markerless human motion capture~\cite{mehta2017vnect,mehta2020xnect,metro}, sports analysis, and the movie and entertainment industry.
A particular difficulty occurs when the human motion is fast, or low light conditions demand longer camera exposure times, which can both lead to blurry images from which it is significantly harder to estimate the human pose.

The main goal of our method is accurate 3D human pose tracking from substantially blurred images or videos.
Hence, it is related to both human pose estimation and image deblurring methods.
On the one hand, while there is a variety of methods that address 3D human pose estimation from RGB or RGB-D images, there is no method that is designed to handle substantially blurred images.
Moreover, none of the human pose estimation methods is able to estimate human pose at sub-frame accuracy.
On the other hand, there is a large amount of methods that aim at deblurring images and videos, but they mostly only assume simplified scenarios, \eg without out-of-image-plane object rotations, or only for rigidly moving objects~\cite{sfb,mfb}.
So far, human pose estimation and image deblurring has not been studied jointly.
Also, there is no public dataset to evaluate such task since none of standard datasets for human pose estimation include significant amounts of motion blur.

\newcommand{\addimgT}[1]{\includegraphics[width=0.16\linewidth]{#1.png}}
\newcommand{\addimgfT}[1]{\framebox{\includegraphics[width=0.16\linewidth]{#1.png}}}
\newcommand{\addimgrT}[1]{\framebox{\includegraphics[trim={5cm 5cm 5cm 5cm},clip,width=0.19\linewidth,height=0.19\linewidth]{#1}}}
\newcommand{\addimgBig}[1]{\includegraphics[width=0.22\linewidth]{image/teaser/#1.png}}
\newcommand{\AlgName}[2]{\rotatebox{90}{\phantom{#2} \scriptsize #1}}

\newcommand{\makeOneRowTeaser}[2]{ 
\addimgT{#10} & \addimgT{#12}  & \addimgT{#14} & \addimgT{#16} & \addimgT{#17} \\
}

\newcommand{\AlgNameTeaser}[2]{\multirow{1}{*}[#2]{\rotatebox{90}{\tiny #1 \phantom{Iy}}}}
\newcommand{\AlgNameTeaserR}[2]{\multirow{1}{*}[#2]{\rotatebox{-90}{\tiny #1 \phantom{Iy}}}}

\newcommand{\makeRowTeaser}[1]{ 
 \addimgT{#1_ib} & \makeOneRowTeaser{#1_hfb_r}  \\
\addimgT{#1_bg} &  \makeOneRowTeaser{#1_hfb_s} \\
\addlinespace[0.1em]
}

\newcommand{\makeOneRowTeaserReverse}[2]{ 
\addimgT{#17} & \addimgT{#16}  & \addimgT{#14} & \addimgT{#12} & \addimgT{#10} \\
}

\newcommand{\makeRowTeaserReverse}[1]{ 
 \addimgT{#1_ib} & \makeOneRowTeaserReverse{#1_hfb_r}  \\
\addimgT{#1_bg} &  \makeOneRowTeaserReverse{#1_hfb_s} \\
\addlinespace[0.1em]
}

\newcommand{\makenewteaser}[1]{
\AlgNameTeaser{Blur}{4em}& \addimgT{#1im_crop} & \addimgT{#1sharp_0}& \addimgT{#1sharp_1} & \addimgT{#1sharp_2}& \addimgT{#1sharp_3} & \addimgT{#1sharp_4} &\AlgNameTeaserR{GT Frames}{5em}\\ \addlinespace[0.2em]
 \cline{3-7}
\addlinespace[0.2em]
\AlgNameTeaser{Background}{5em} &\addimgT{#1bgr_crop} & \addimgT{#1rendered_0}& \addimgT{#1rendered_1} & \addimgT{#1rendered_2}& \addimgT{#1rendered_3} &\addimgT{#1rendered_4} & \AlgNameTeaserR{Rendered}{5em} \\
       \addlinespace[0.4em]       
\AlgNameTeaser{Matting~\cite{BGMv2}}{5em}&\addimgT{#1labeled_c} & \addimgfT{#1novel_view_0}& \addimgfT{#1novel_view_1} & \addimgfT{#1novel_view_2}& \addimgfT{#1novel_view_3} & \addimgfT{#1novel_view_4} &\AlgNameTeaserR{Novel View}{5em}\\
        
                \addlinespace[0.3em]}
\begin{figure}
\centering
\scriptsize
\setlength{\tabcolsep}{0.05em} 
\renewcommand{\arraystretch}{0.2} 
\begin{tabular}{@{}c@{\hskip 0.01em}c@{\hskip 0.5em}cccccc@{}}

\makenewteaser{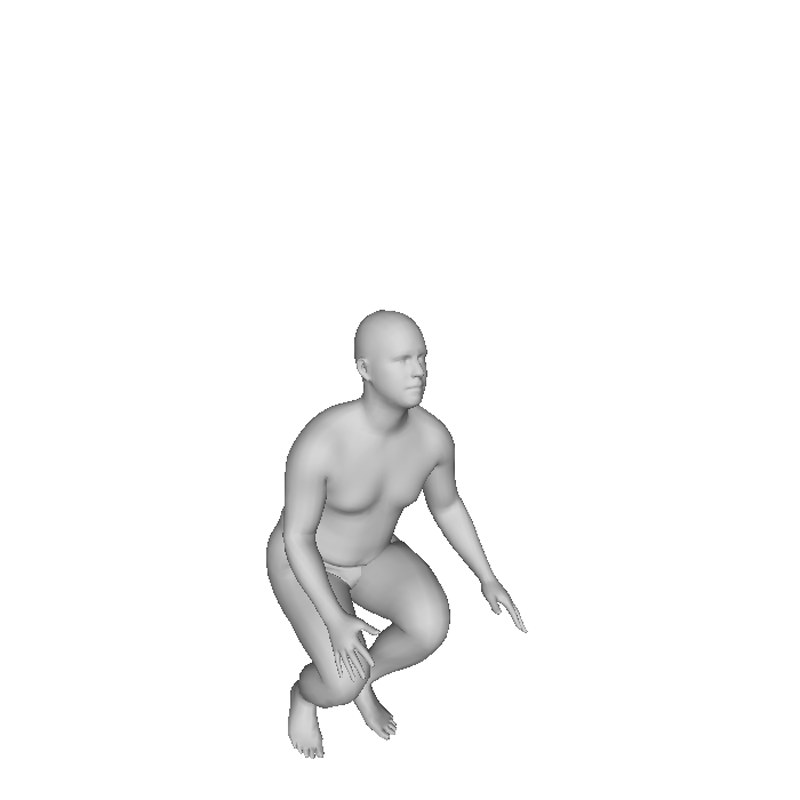}
&\MakeArrowVar{1} &  \MakeArrowVar{5} \\\addlinespace[0.1em]
& Inputs &  \multicolumn{5}{c}{Outputs (Estimated sub-frame pose) }
\end{tabular}
\vspace{-0.2em}
\caption{\textbf{Human from Blur (HfB) on a real-world sequence.} Given a blurry image with human motion and the corresponding background, HfB recovers the human shape and sub-frame motion. We visualize sub-frame human pose and show the reconstructed mesh from a novel view. 
}
\label{fig:teaser}
\end{figure}

\begin{figure*}[t]
\centering
\includegraphics[width=\linewidth]{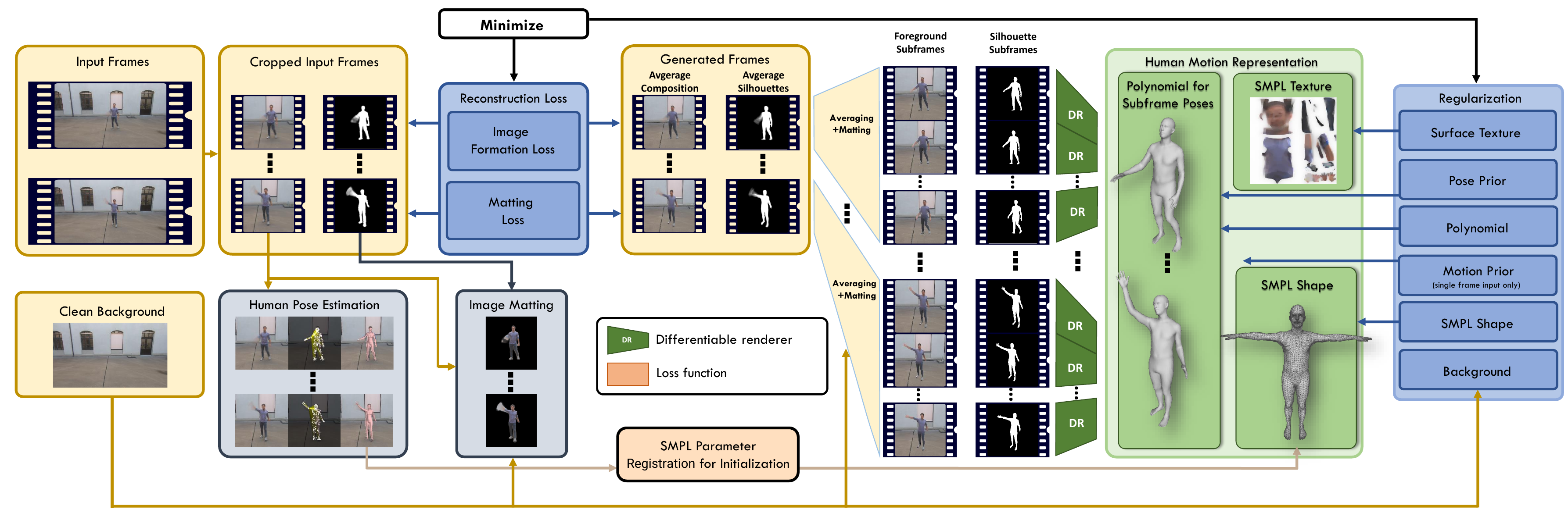}\\
\caption{\textbf{Method overview.} The input to our method are a single or multiple blurry frames of a human (left), and the output is a 3D representation of a human and its sub-frame motion over time (right). \textbf{From Right to Left:} Starting from the human motion representation, our model can be seen as generative model. For a desired set of frames and sub-frames, we can render sub-frame appearances and corresponding silhouettes. Then, the sub-frames are averaged to generate blurry frames and blurry silhouettes (alpha channel), which are composed with the known background to generate the input image according to~\eqref{eq:formation}.
The central part of our method is the image reconstruction loss which compares the generated images with the actual input images. In order to solve for the human motion estimation, the reconstruction loss is backpropagated through the entire differentiable pipeline. The human pose estimation uses a traditional method~\cite{metro} to initialize the optimization, and the image matting is precomputated~\cite{BGMv2} for the matting loss.
}
\label{fig:pipeline}
\end{figure*}

We propose the first method that recovers human pose at sub-frame accuracy from blurry inputs, even from a single blurry image (Fig.~\ref{fig:teaser}).

We make the following contributions:
\begin{enumerate}[itemsep=0.0pt,topsep=3pt,leftmargin=*,label=\textbf{(\arabic*)}]
    \item We present the first method for human pose estimation from substantially blurred images that recovers sub-frame accurate poses as well as texture and body shape.
    \item We generate a synthetic dataset and collected real-world motion-blurred data of humans for evaluation purposes. We further propose corresponding evaluation metrics to assess and compare to future methods.
    \item The proposed method only relies on test-time optimization and is learning-free, apart from the initialization and the motion prior, which is only needed for the single-frame case.
    Hence, our method does not require large amounts of annotated training data.
\end{enumerate}

\section{Related work}

The proposed method is at the intersection of human pose tracking and image/video deblurring.

\boldparagraph{3D Human pose estimation.}
The 3D pose of a human is usually represented as a skeleton of 3D joints \cite{martinez2017simple,mehta2017vnect,mehta2020xnect,zhou2016sparseness,sun2018integral}.
In order to obtain more fine-grained representations of the human body, parametric body models such as SCAPE~\cite{anguelov2005scape} or the SMPL family~\cite{smpl, SMPL-X:2019,osman2020star} have been introduced to capture the 3D body pose.
Iterative optimization-based approaches have been leveraged for model-based human pose estimation. \cite{guan2009estimating,sigal2008combined,hasler2010multilinear,bogo2016keep,SMPL-X:2019} proposed to estimate the parameters of the human model by leveraging silhouettes or 2D keypoints. 
On the other hand, direct parameter regression via neural networks has been explored \cite{kanazawa2018end,tung2017self,tan2018indirect,omran2018neural,guler2019holopose,varol2017learning,xu2019denserac,zheng2019deephuman, li2021hybrik, ROMP,metro, kocabas2019vibe}. 
Given a single RGB image, a deep network is used to regress the human model parameters. 
There is another line of work that combines the advantages of both optimization and regression to fit the SMPL body~\cite{kolotouros2019spin, song2020lgd}.
Although there have been significant advances of human pose estimation from monocular images or videos, a method which is able to deal with blurry input is still missing.

\boldparagraph{Image and video deblurring.}
A large amount of methods have studied generic image~\cite{Kupyn_2018_CVPR,Kupyn_2019_ICCV} and video deblurring,~\eg~\cite{Jin_2019_CVPR,Pan_2020_CVPR,zhong2020efficient,BIN,zhou2019stfan, Chi_2021_CVPR, Li_2021_CVPR, Zhang_2020_CVPR, Suin_2020_CVPR, Kaufman_2020_CVPR, tbd3d, defmo}.
Some attempts to specialize on deblurring depicted humans have already been made.
For instance, \cite{chrysos2019motion} focuses only on deblurring human faces.
Closely related to our problem setting, \cite{lumentut2020human} addresses deblurring of human motion using an adversarial approach, which focuses on image deblurring rather than pose estimation, and it does not recover at sub-frame accuracy.
The follow-up method~\cite{lumentut2021human} generalizes to joint human motion and scene deblurring with a similar methodology, but the sub-frame poses are never recovered. 

The proposed method is partially inspired by Shape from Blur (SfB)~\cite{sfb}, which uses a similar test-time optimization to recover 3D shape and sub-frame motion of simple rigid objects with spherical topology from a single blurry image with a given background.
Motion from Blur (MfB)~\cite{mfb} extends SfB to multiple video frames.
There is also a related Animation from Blur method~\cite{zhong2022animation}, but it assumes a motion guidance is provided as an additional input.

\section{Method}

The inputs to our method are an image $\im$ with the blurred human and the corresponding clean background image $\bgr$.
The desired output is a human shape parameter $\shape$, texture image $\texture$, and three functions representing sub-frame human motion that depend on a timestamp $\tm$. 
This timestamp represents the sub-frame time interval and is defined between $1$ and $N$, where $N$ is the desired number of sub-frames.
Effectively, it means that we generate a temporal super-resolution or a short video with $N$ frames out of each single input frame.
Those three functions are human body translation $\tran$, rotation $\rot$, and sub-frame human pose $\pose$ that represents joint rotation.
They are all represented by a set of low-degree polynomials, where translations and rotations have each four degrees of freedom (direction with distance and axis with angle).
This polynomial representation generates poses in a strict chronological order and is continuous, differentiable, and can be easily initialized with a given initial pose (Sec.~\ref{sec:init}).
Human pose and shape representations follow the SMPL human model~\cite{smpl}.
The texture image $\texture$ is mapped using a fixed UV mapping from SMPL.

As the first step, we generate the human SMPL mesh $\mesh$ at timestamp $\tm$ with a given pose, shape, and texture parameters (Fig.~\ref{fig:pipeline}).
Then, we move the whole mesh $\mesh$ by translation $\tran$ and rotation $\rot$ given by motion function $\motion$:
\begin{equation}
  \mesh = \motion \big( \text{SMPL}(\pose, \shape, \texture), \tran, \rot \big) \enspace.
\end{equation}
To render the sub-frame silhouette and appearance of the mesh, we use Differentiable Interpolation-based Renderer (DIB-R)~\cite{dibr}. 
This differentiable rendering provides two outputs.
The first one is appearance rendering $\renf(\mesh)$ that outputs projected human appearance.
The second one is silhouette rendering $\rens(\mesh)$ that outputs projected human silhouette.
In this work, we assume a static camera.

\boldparagraph{Image formation model.}
%
Given all previously defined parameters, we can finally define the image formation model.
It follows a standard alpha matting approach:
\begin{equation}
  \hat{\im} = \underbrace{\Big(\!1-\frac{1}{N}\!\sum_{i=1} ^N \rens(\mesh)\Big)}_\text{Inverse alpha channel} \!\cdot
  \vertarrowbox[2.9ex]{B}{Background} +
\underbrace{\frac{1}{N}\!\sum_{i=1}^N \rens(\mesh)\!\cdot\! \renf (\mesh)}_\text{Blurred foreground (human)} \, .
\label{eq:formation}
\end{equation}

The generated image $\hat{\im}$ consists of the background image, scaled down by the inverse alpha channel, and the blurred foreground human body.
The alpha channel is modeled by averaging all projected sub-frame silhouettes.

\subsection{Loss terms}
%
The key components of our method are the image formation loss and the matting loss. 
The image formation loss forces the reconstructed image to be as close as possible to the input image.
The matting loss favors silhouettes that are consistent with the initially estimated alpha channel.
The other losses are auxiliary and regularization terms that make the optimization easier and refine the final results.  

\boldparagraph{Image formation loss.} 
This loss measures the input image reconstruction according to the image formation model~\eqref{eq:formation}. 
We compute the mean squared error between the observed input image and our reconstruction as:

\begin{equation}
  \mathcal{L}_I= | \im - \hat{\im} |_2  \enspace.
  \label{eq:lossi}
\end{equation}

\boldparagraph{Matting loss.}
If the image formation loss~\eqref{eq:lossi} is the sole loss to be minimized, the optimization becomes extremely difficult. 
Experimentally, such optimization is ambiguous and mostly results in an undesired local minimum.
Therefore, we further impose a loss on our approximated rendered alpha channel, which is computed as the average of sub-frame silhouettes, $\alpha_\text{target} = \frac{1}{N}\sum_{i=1}^N \rens(\mesh)$, according to the image formation model~\eqref{eq:formation}. 
The initial alpha channel $\alpmat$ is estimated using a pre-trained Background-Matting-V2~\cite{BGMv2} model, based on the input blurry image and the corresponding background.
Finally, the matting loss computes the intersection over union between our rendered alpha channel from averaging and the one from~\cite{BGMv2}:
\begin{equation}
  \mathcal{L}_\alpha = 1 - \frac{| \min(\alpha_\text{in},\alpha_\text{target}) |_1}{| \max(\alpha_\text{in},\alpha_\text{target}) |_1}  \enspace ,
  \label{eq:lossl}
\end{equation}
where the intersection over union for non-binary inputs is a ratio between the sum of pixel-wise $\min$ and $\max$ operators.

\boldparagraph{Surface texture smoothness.}
The UV texture map from SMPL contains many non-overlapping regions (see Fig.~\ref{fig:real}, HfB row), and the correct neighborhoods are not properly defined.
Therefore, the commonly used total variation loss for texture smoothness~\cite{mfb} cannot be directly applied in this case since it will propagate the color of the void area.

To address this issue, we propose a surface texture smoothness term that accounts for the mesh faces neighborhood.
For a given texture pixel $p_k$ and its $8$ surrounding neighboring pixels $p_j \in \mathcal{N}(p_k)$, we pick those ones that are neighbors in the mesh ($\mathbf{c}_{k,j}=1$), \ie they belong to adjacent triangular faces, and that are visible in at least one of the sub-frames ($\mathbf{v}_{j} = 1$).
Then, we compute the cosine value between the face normal $n_k$ of the current pixel and the face normal $n_j$ of its chosen neighbors.
The introduction of the cosine of face normals takes into account the mesh geometry, \ie the texture should be smoother on flat surfaces.
Then, the surface texture smoothness is expressed as a weighted sum of absolute differences in RGB pixels:
\begin{equation}
\label{eq:losss}
  \mathcal{L}_S= \frac{1}{8 |\texture|}\!\sum_{p_k \in \texture} \sum_{p_j \in \mathcal{N}(p_k)} \!\!\!\!\mathbf{c}_{k,j} \mathbf{v}_{j}\cos \angle(n_k,n_j)|p_k - p_j|_1  \, .
\end{equation}
\renewcommand{\addimgT}[1]{\framebox{\includegraphics[width=0.07\linewidth]{#1.png}}}
\newcommand{\addimgcropreal}[1]{\framebox{\includegraphics[trim={0.4cm 1cm 1.0cm 0.0cm},clip,width=0.07\linewidth,height=0.07\linewidth]{#1}}}


\newcommand{\makeOneRowReal}[1]{\addimgT{#10} & \addimgT{#11}  & \addimgT{#12} & \addimgT{#13} & \addimgT{#14} & \addimgT{#15} & \addimgT{#16} & \addimgT{#17} }

\newcommand{\makeOneRowSfBReal}[1]{\addimgcropreal{#10} & \addimgT{#11}  & \addimgT{#12} & \addimgT{#13} & \addimgT{#14} & \addimgT{#15} & \addimgT{#16} & \addimgT{#17} }

\newcommand{\makeRowSingleReal}[1]{ 
 & & 
 \AlgName{SfB~\cite{sfb}}{a} & 
 \makeOneRowSfBReal{#1_sfb_} & \addimgT{#1_sfb_r} &
 \addimgT{#1_sfb_tex} \\
\addimgT{#1_ib} & \addimgT{#1_ci} &
\AlgName{HfB (ours)}{} & 
\makeOneRowReal{#1_r} & \addimgT{#1_ir} &
\addimgT{#1__tex} \\
 \addimgT{#1_bg} & \addimgT{#1_lb} &
 \AlgName{$\renf$}{Aa} & 
 \makeOneRowReal{#1_a} &   \addimgT{#1_mean_app}\\
& & \AlgName{$\rens$}{Aa} & \makeOneRowReal{#1_s} & \addimgT{#1_av}  \\
}

\newcommand{\makeOneRowRealReverse}[1]{\addimgT{#17} &  \addimgT{#16} & \addimgT{#15}  & \addimgT{#14}   & \addimgT{#13} & \addimgT{#12} & \addimgT{#11} & \addimgT{#10}  }

\newcommand{\makeRowSingleRealReverse}[1]{ 
 & & 
 \AlgName{SfB~\cite{sfb}}{a} & 
 \makeOneRowRealReverse{#1_sfb_} & \addimgT{#1_sfb_r} &
 \addimgT{#1_sfb_tex} \\
\addimgT{#1_ib} & \addimgT{#1_ci} &
\AlgName{HfB (ours)}{} & 
\makeOneRowRealReverse{#1_r} & \addimgT{#1_ir} &
\addimgT{#1__tex} \\
 \addimgT{#1_bg} & \addimgT{#1_lb} &
 \AlgName{$\renf$}{Aa} & 
 \makeOneRowRealReverse{#1_a} &   \addimgT{#1_mean_app}\\
& & \AlgName{$\rens$}{Aa} & \makeOneRowRealReverse{#1_s} & \addimgT{#1_av}  \\
}

\begin{figure*}
\centering
\small
\setlength{\tabcolsep}{0.05em} 
\renewcommand{\arraystretch}{0.4} 
\begin{tabular}{@{}cc@{\hskip 0.8em}ccccccccc@{\hskip 0.4em}c@{\hskip 0.4em}c@{}}
\makeRowSingleReal{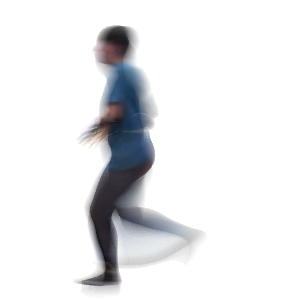} \addlinespace[0.5em]
\makeRowSingleReal{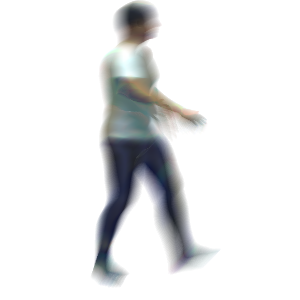}
Inputs & Initialization  & \multicolumn{9}{c}{Sub-frames} & Averages & Est. texture \\

\end{tabular}
\caption{\textbf{Results on real data that we captured.}
The proposed method significantly outperforms SfB~\cite{sfb} and provides plausible human shape and pose reconstructions. \textbf{Left}: initialization of human pose from METRO~\cite{metro} (top) and alpha channel from~\cite{BGMv2} (bottom).}
\label{fig:real}
\end{figure*}

\boldparagraph{Pose prior loss.}
We import the pose prior loss from SMPLify-X \cite{SMPL-X:2019}.
This prior scores how feasible are the estimated pose parameters $\pose$:
\begin{equation}
\label{eq:lossp}
  \mathcal{L}_P = \frac{1}{N}\sum_{i=1}^N \text{prior}(\pose)  \enspace.
\end{equation}

\boldparagraph{SMPL shape regularization.}
We add norm regularization on the SMPL shape parameter $\shape$ to avoid irregular human body shape as used in SMPL~\cite{smpl}:
\begin{equation}
\label{eq:lossbeta}
  \mathcal{L}_\beta = | \beta |_2^2  \enspace.
\end{equation}

\boldparagraph{Polynomial regularization.}
The polynomial coefficients of the pose, translation, and rotation could be serialized into a matrix $\coeff \in \mathbb{R}^{4d \times (J+2)}$, where $d$ is the degree of the polynomial, and $J$ is the number of joints in the SMPL model. 
The whole body translation and rotation are already incorporated into matrix $\coeff$, thus we have $4 (J+2)$ polynomials of $d$ degree.
Since rotations, translations, and joint poses have 4 degrees of freedom each, we have a separate polynomial for each degree of freedom.
We apply both the L1-norm and the Frobenius norm on the polynomial coefficients:
\begin{equation}
\label{eq:lossc}
    \mathcal{L}_C = | \coeff |_1 + | \coeff |_F \enspace.
\end{equation}
The intention of adding this regularization is to avoid extreme joint movement.

\boldparagraph{Background regularization.}
We assume that the human texture is sufficiently distinct from the background. 
This is enforced by the difference between the projected
object appearance and the background:
\begin{equation}
\label{eq:lossb}
    \mathcal{L}_B = \frac{1}{N}\sum_{\tm=1} \frac{1}{|\bgr-\renf(\mesh)|+\epsilon} 
    \quad\text{with}\;\epsilon = 10^{-6} \enspace.
\end{equation}

\setlength{\fboxsep}{0pt}
\renewcommand{\addimgT}[1]{\framebox{\includegraphics[width=0.073\linewidth]{#1.png}}}

\newcommand{\addimgBigsingle}[1]{\includegraphics[width=0.22\linewidth]{image/single_compare_2/#1.png}}

\newcommand{\AlgNametwo}[2]{\rotatebox{90}{\phantom{#2} \tiny  #1}}

\newcommand{\addinputs}[1]{\raisebox{3em}{\multirow{3}{*}{ \begin{tabular}{c@{\hskip 0.4em}c}
\framebox{\includegraphics[width=0.2\linewidth]{#1im_crop.png}}\\ \addlinespace[0.2em]
Input (blur rate 0.54)  \end{tabular} }}}
\newcommand{\addmetro}[1]{\raisebox{3em}{\multirow{3}{*}{ \begin{tabular}{c@{\hskip 0.4em}c}
\framebox{\includegraphics[width=0.2\linewidth]{#1metro_pred_0.png}}\\ \addlinespace[0.2em]
METRO \cite{metro}\end{tabular} } }}
  
\newcommand{\addimgcropsyn}[1]{\framebox{\includegraphics[trim={0.4cm 1cm 1.0cm 0.4cm},clip,width=0.073\linewidth,height=0.073\linewidth]{#1.png}}}
\newcommand{\addimgcropmean}[1]{\framebox{\includegraphics[trim={0.2cm 1cm 0.9cm 0.4cm},clip,width=0.073\linewidth,height=0.073\linewidth]{#1.png}}}
\newcommand{\AlgNameMu}[2]{\multirow{1}{*}[#2]{\rotatebox{90}{\scriptsize #1 \phantom{Iy}}}}
\newcommand{\makeRowSingle}[1]{ 
   \addinputs{#1}      & \AlgNameMu{GT}{10pt}  & \addimgT{#1gt_body_0} & \addimgT{#1gt_body_1} & \addimgT{#1gt_body_2} & \addimgT{#1gt_body_3} & \addimgT{#1gt_body_4} & \addimgT{#1gt_body_5} &  \addimgT{#1gt_body_6}   \\
  &       & \addimgT{#1gt_s_0} & \addimgT{#1gt_s_1} & \addimgT{#1gt_s_2} & \addimgT{#1gt_s_3} & \addimgT{#1gt_s_4}&  \addimgT{#1gt_s_5}&  \addimgT{#1gt_s_6} \\
 \addlinespace[0.2em]
                 & \AlgNameMu{HfB (ours)}{37pt}        & \addimgT{#1hfb_body_0} & \addimgT{#1hfb_body_1} & \addimgT{#1hfb_body_2} & \addimgT{#1hfb_body_3} & \addimgT{#1hfb_body_4}&  \addimgT{#1hfb_body_5}&  \addimgT{#1hfb_body_6} \\
 \addlinespace[0.2em]
  \addmetro{#1}  &  \AlgNameMu{AfB~\cite{zhong2022animation}+\cite{metro}}{30pt}       & \addimgT{#1afb_s_0} & \addimgT{#1afb_s_6} & \addimgT{#1afb_s_5} & \addimgT{#1afb_s_4} & \addimgT{#1afb_s_3}&  \addimgT{#1afb_s_2}&  \addimgT{#1afb_s_1} \\
                 &   & \addimgT{#1afb_a_0} & \addimgT{#1afb_a_6} & \addimgT{#1afb_a_5} & \addimgT{#1afb_a_4} & \addimgT{#1afb_a_3} & \addimgT{#1afb_a_2} &  \addimgT{#1afb_a_1}  \\ 
 \addlinespace[0.1em]                  
                 &  \AlgNameMu{Jin~\etal~\cite{Jin_2018_CVPR} +\cite{metro}}{35pt}       & \addimgT{#1jin_s_0} & \addimgT{#1jin_s_1} & \addimgT{#1jin_s_2} & \addimgT{#1jin_s_3} & \addimgT{#1jin_s_4}&  \addimgT{#1jin_s_5}&  \addimgT{#1jin_s_6} \\
                 &    & \addimgT{#1jin_a_0} & \addimgT{#1jin_a_1} & \addimgT{#1jin_a_2} & \addimgT{#1jin_a_3} & \addimgT{#1jin_a_4} & \addimgT{#1jin_a_5} &  \addimgT{#1jin_a_6}  \\ 
}
\begin{figure*}
\centering
\small
\setlength{\tabcolsep}{0.05em} 
\renewcommand{\arraystretch}{0.2} 
\begin{tabular}{@{}c@{\hskip 0.8em}cccccccc@{}}
\makeRowSingle{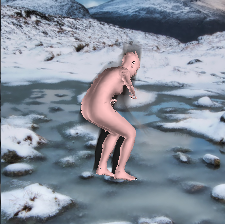} 
\addlinespace[0.2em]
  & & \multicolumn{7}{c}{Sub-frames}\\
\end{tabular}
\caption{\textbf{Comparison on synthetic data.}
Given a blurry image with human motion, sub-frame human poses generated by HfB are consistent, whereas for learning-based temporal super-resolution methods~\cite{Jin_2018_CVPR,zhong2022animation} (with METRO~\cite{metro} human poses on sub-frames) the poses are not consistent, \eg motion of the right arm. 
We also visualize the raw METRO~\cite{metro} pose prediction on the input blurry image. }
\label{fig:synth_sfb}
\end{figure*}

%

\boldparagraph{Adversarial short motion prior.} \label{para:amp}
Since the human body consists of multiple joints, there exists a significant amount of ambiguity in case of a single input blurry image.
The ambiguity comes mainly from the unknown motion direction.
In fact, both the forward and the backward directions provides the same blurry image according to the image formation model~\eqref{eq:formation}.
Potentially, there are exponentially many motion directions for each joint that lead to the same input.
And it is infeasible to estimate the correct direction directly from a single image without any additional priors.
Otherwise, the choice of motion direction will be arbitrary. Many prior studies~\cite{SMPL-X:2019,kocabas2019vibe} offer motion priors, but they are not suitable for our setting.

To address this problem we propose the adversarial motion prior to recognize wrong (reversed) motion of joints. 
Based on our polynomial motion representation, we propose an adversarial model that could supervise on the polynomial coefficients $\coeff$.
The model is inspired by the image in-painting methods~\cite{pathakCVPR16context,inbook}.

Our adversarial model consists of two components: a discriminator $D$ that generates a binary indicator function to identify unrealistic entries in the coefficients $\coeff$, and a correction-generator $G$ that predicts realistic polynomial coefficients from the given polynomial coefficients $\coeff$ and the indicator function $\indic$.

The training data are sampled from the AMASS dataset~\cite{AMASS:2019} (CMU~\cite{cmuWEB} and ACCAD~\cite{ACCAD}).
The training is supervised jointly by four loss terms.
The discriminator loss is the binary cross entropy loss, which is applied to the indicator function predicted by the discriminator and compared to the ground truth. 
The generator loss, specifically the reconstruction loss, comprises three terms.
The first one is L1 loss between the coefficient matrix predicted by the correction- generator and the ground truth. 
The second is L2 loss between the reconstructed pose and the ground truth pose.
The last one is the mean per joint position error (MPJPE)~\cite{song2020lgd} between the reconstructed joint positions and the ground truth SMPL joint positions. 

This adversarial model is pre-trained as mentioned above and is fixed during optimization. 
In case of a single input blurry image, the adversarial motion prior is incorporated into the optimization as the L1 loss between the generator output and the polynomial coefficients $\coeff$:
\begin{equation}
\label{eq:lossla}
    \mathcal{L}_A =|G(D(\coeff), \coeff) - \coeff|_1 \enspace.
\end{equation}

\boldparagraph{Joint loss.}
%
The final loss is a weighted sum of all previously defined losses:
\begin{equation}
\begin{split}
    \mathbf{\mathcal{L}} = & 
      w_I \mathcal{L}_I + w_\alpha \mathcal{L}_\alpha +
      w_S \mathcal{L}_S + w_P\mathcal{L}_P + \\ &
      w_\beta \mathcal{L}_\beta + w_C \mathcal{L}_C +
      w_B \mathcal{L}_B  +w_A \mathcal{L}_A  \enspace.
\end{split}\label{eq:lossj}
\end{equation}

\newcommand{\mySkip}{\hskip 0.25em}
\newcommand{\mpjpe}{\scriptsize MPJPE\!$\downarrow$\!\!}
\newcommand{\iou}{\scriptsize IoU\!$\uparrow$\!\!}
\begin{table*}[!htbp]
\scriptsize
\setlength{\tabcolsep}{1.8pt}
\renewcommand{\arraystretch}{1.1} 
\begin{tabular}{lc@{ \mySkip}cc@{ \mySkip}cc@{ \mySkip}cc@{ \mySkip}cc@{ \mySkip}cc@{ \mySkip}cc@{ \mySkip}cc@{ \mySkip}cc@{ \mySkip}cc@{ \mySkip}c}
\toprule
\begin{tabular}[c]{@{}l@{}}blur rate int. \\ \#images\end{tabular} & \multicolumn{2}{c}{\begin{tabular}[c]{@{}c@{}}[0.05, 0.1]\\ 123\end{tabular}} & \multicolumn{2}{c}{\begin{tabular}[c]{@{}c@{}}[0.1, 0.2]\\ 206\end{tabular}} & \multicolumn{2}{c}{\begin{tabular}[c]{@{}c@{}}[0.2, 0.3]\\ 154\end{tabular}} & \multicolumn{2}{c}{\begin{tabular}[c]{@{}c@{}}[0.3, 0.4]\\ 139\end{tabular}} & \multicolumn{2}{c}{\begin{tabular}[c]{@{}c@{}}[0.4, 0.5]\\ 106\end{tabular}} & \multicolumn{2}{c}{\begin{tabular}[c]{@{}c@{}}[0.5, 0.6]\\ 108\end{tabular}} & \multicolumn{2}{c}{\begin{tabular}[c]{@{}c@{}}[0.6, 0.7]\\ 81\end{tabular}}  & \multicolumn{2}{c}{\begin{tabular}[c]{@{}c@{}}[0.7, 0.8]\\ 79\end{tabular}}  & \multicolumn{2}{c}{\begin{tabular}[c]{@{}c@{}}[0.8, 0.9]\\ 76\end{tabular}}  & \multicolumn{2}{c}{\begin{tabular}[c]{@{}c@{}}[0.9, 1.1]\\ 52\end{tabular}}  \\ \cmidrule(r){1-1} \cmidrule(lr){2-3} \cmidrule(lr){4-5} \cmidrule(lr){6-7} \cmidrule(lr){8-9} \cmidrule(lr){10-11} \cmidrule(lr){12-13} \cmidrule(lr){14-15} \cmidrule(lr){16-17}  \cmidrule(lr){18-19}  \cmidrule(lr){20-21}  
Method\!\!    & \mpjpe & \iou & \mpjpe & \iou & \mpjpe & \iou & \mpjpe & \iou & \mpjpe & \iou  & \mpjpe & \iou & \mpjpe & \iou & \mpjpe & \iou & \mpjpe & \iou & \mpjpe & \iou  \\ \cmidrule(r){1-1} \cmidrule(lr){2-3} \cmidrule(lr){4-5} \cmidrule(lr){6-7} \cmidrule(lr){8-9} \cmidrule(lr){10-11} \cmidrule(lr){12-13} \cmidrule(lr){14-15} \cmidrule(lr){16-17}  \cmidrule(lr){18-19}  \cmidrule(lr){20-21}  
SfB\cite{sfb}       & N.A. & 0.498    &N.A. & 0.487  &N.A.  &0.493 &N.A.  &0.455 &N.A. &0.428 & N.A. & 0.408    &N.A.  & 0.409 &N.A.  &0.397 &N.A.  &0.378 &N.A. &0.363
\\
METRO \cite{metro}  & 70.0         & 0.632          & 71.9         & 0.637          & 84.1          & 0.615          & 101.2         & 0.590         &  121.3         & 0.540          &  121.4           & 0.500          &   132.8          & 0.489            &     143.4               & 0.428           &  147.1         & 0.421            & 146.4    & 0.423   \\
HfB w/o AMP         & 65.1         & 0.829          & 65.7         & 0.813          & 68.7          & 0.803          & 80.6          & 0.785         &  98.5          & 0.763          &  107.2           & 0.739          &   115.1          & 0.731            &     129.0       &  \textbf{0.685}   &\textbf{ 121.7} &\textbf{ 0.670}   &\textbf{ 138.9}   & \textbf{0.645} \\
HfB (ours) &\textbf{ 56.3} & \textbf{0.859} &\textbf{ 59.4}& \textbf{0.837} & \textbf{66.4} & \textbf{0.820} &\textbf{ 78.3} & \textbf{0.805} & \textbf{89.0} & \textbf{0.775} & \textbf{ 101.1}  & \textbf{0.743 }  & \textbf{110.8} & \textbf{0.734 }          & \textbf{128.7}           & 0.682            & 124.6          & 0.659            & 140.5    & 0.633  \\ \bottomrule 

\end{tabular}
\vspace{0.3em}
 \caption{\textbf{Single-frame evaluation for different blur rates on BT-AMASS dataset.} The proposed method outperforms SfB \cite{sfb} (no human pose output) and METRO \cite{metro}, which we also use for initialization. Our method improves significantly over the initialization. The proposed AMP prior (Sec.~\ref{para:amp}) improves results only slightly, and even becomes harmful for higher blur rates due to more ambiguity.}
  \label{tab:single_results}
\end{table*}

\renewcommand{\mySkip}{\hskip 0.25em}

\begin{table*}[!htbp]
\scriptsize
 \setlength{\tabcolsep}{1.8pt} 
\renewcommand{\arraystretch}{1.1} 
\begin{tabular}{lc@{ \mySkip}cc@{ \mySkip}cc@{ \mySkip}cc@{ \mySkip}cc@{ \mySkip}cc@{ \mySkip}cc@{ \mySkip}cc@{ \mySkip}cc@{ \mySkip}cc@{ \mySkip}c}
\toprule
\begin{tabular}[c]{@{}l@{}}blur rate int. \\ \#videos\end{tabular} & \multicolumn{2}{c}{\begin{tabular}[c]{@{}c@{}}[0.05, 0.1]\\ 35\end{tabular}} & \multicolumn{2}{c}{\begin{tabular}[c]{@{}c@{}}[0.1, 0.2]\\ 39\end{tabular}} & \multicolumn{2}{c}{\begin{tabular}[c]{@{}c@{}}[0.2, 0.3]\\ 47\end{tabular}} & \multicolumn{2}{c}{\begin{tabular}[c]{@{}c@{}}[0.3, 0.4]\\ 42\end{tabular}} & \multicolumn{2}{c}{\begin{tabular}[c]{@{}c@{}}[0.4, 0.5]\\ 38\end{tabular}} & \multicolumn{2}{c}{\begin{tabular}[c]{@{}c@{}}[0.5, 0.6]\\ 32 \end{tabular}}  & \multicolumn{2}{c}{\begin{tabular}[c]{@{}c@{}}[0.6, 0.7]\\ 25\end{tabular}}  & \multicolumn{2}{c}{\begin{tabular}[c]{@{}c@{}}[0.7, 0.8]\\ 19\end{tabular}}  & \multicolumn{2}{c}{\begin{tabular}[c]{@{}c@{}}[0.8, 0.9]\\ 13\end{tabular}}  & \multicolumn{2}{c}{\begin{tabular}[c]{@{}c@{}}[0.9, 1.1]\\ 15\end{tabular}}  \\ \cmidrule(r){1-1} \cmidrule(lr){2-3} \cmidrule(lr){4-5} \cmidrule(lr){6-7} \cmidrule(lr){8-9} \cmidrule(lr){10-11} \cmidrule(lr){12-13} \cmidrule(lr){14-15} \cmidrule(lr){16-17}  \cmidrule(lr){18-19}  \cmidrule(lr){20-21}  
Method\!\!    & \mpjpe & \iou & \mpjpe & \iou & \mpjpe & \iou & \mpjpe & \iou & \mpjpe & \iou  & \mpjpe & \iou & \mpjpe & \iou & \mpjpe & \iou & \mpjpe & \iou & \mpjpe & \iou  \\ \cmidrule(r){1-1} \cmidrule(lr){2-3} \cmidrule(lr){4-5} \cmidrule(lr){6-7} \cmidrule(lr){8-9} \cmidrule(lr){10-11} \cmidrule(lr){12-13} \cmidrule(lr){14-15} \cmidrule(lr){16-17}  \cmidrule(lr){18-19}  \cmidrule(lr){20-21}  
MfB~\cite{mfb}                 & N.A.           & 0.513              &N.A.          & 0.547            &N.A.            &0.528             &N.A.            &0.509             &N.A.          &0.384           & N.A.            &0.356             &N.A.           &0.342            &N.A.            &0.281           &N.A.             & 0.268          &N.A.           &0.229\\
METRO~\cite{metro}\phantom{A}   & 68.9          & 0.645            & 69.6           &0.689             & 71.6           &0.653            & 67.0           &0.601          &   98.6          & 0.598          &  99.6          & 0.508            & 108.5         & 0.477          &  111.3           & 0.399          &  116.8            & 0.408         & 121.1         & 0.388 \\
 HfB (ours)            & \textbf{ 65.4} & \textbf{0.819}  & \textbf{ 64.4} & \textbf{0.828}  & \textbf{ 69.4} & \textbf{0.823}  & \textbf{ 56.2} &\textbf{ 0.787}  & \textbf{ 77.2} & \textbf{0.779}  & \textbf{ 83.4} &\textbf{ 0.738}  & \textbf{ 102.4} & \textbf{0.766}  & \textbf{ 101.4} & \textbf{0.695} & \textbf{ 109.4} &\textbf{ 0.671}        & \textbf{112.0} & \textbf{0.656}  \\ \bottomrule 

\end{tabular}
\vspace{0em}
\caption{\textbf{Two-frame evaluation for different blur rates on BT-AMASS dataset.} Similarly to Table~\ref{tab:single_results}, HfB outperforms other methods even when there are two input blurry frames. We compare to MfB~\cite{mfb} (multi-frame method) and interpolated poses from METRO~\cite{metro}.   } 
  \label{tab:multi_results}
  \vspace{-1em}
\end{table*}

\subsection{Multiple blurry images  }

Our approach can be extended to  multiple consecutive blurry images in a video.
In this case, the human body shape $\shape$ and texture $\texture$ are assumed to be the same for all input images, while the other parameters, \eg poses, are separate for each frame.
In general, this setting is simpler since there are more constraints from more images. 
Also, there is no more ambiguity in the motion direction of each joint.
Therefore, the adversarial motion prior $\mathcal{L}_A$ is not needed anymore.

For smooth joint motion in consecutive frames, we add a boundary restriction on the joints rotation and position.
For instance, in case of two input blurry images, we add a boundary restriction at the end timestamp $\tm = N$ of the first image and the start timestamp $\tm = 1$ of the second image.
The boundary restriction forces the joints rotation and position and their first order derivatives to be equal at the boundary to preserve the motion continuity, and it is implemented by the L1 loss with a unit weight.
For images with exposure gap, we extend the end timestamp of first image with exposure time $\tau$ (measured in sub-frames) and then apply boundary restriction at $N +\tau$.

\subsection{Optimization}
The joint loss~\eqref{eq:lossj} is minimized using the ADAM optimizer~\cite{adam} for 200 iterations with learning rate $0.01$ on a single 12 GB RTX 2080 Ti graphics card.

\boldparagraph{Initialization.}\label{sec:init}
To initialize our method, we use the METRO~\cite{metro} human pose estimation method, which reconstructs a single human pose from a blurry image reasonably well, albeit without sub-frame accuracy.
We fit the initial body translation, rotation, pose, and shape parameters to the mesh generated from METRO using a SMPL registration model~\cite{bhatnagar2020ipnet}. 
In case of sub-frame translation, rotation, and poses, we initialize the polynomial coefficients at timestamp $\tm = 1$ and all other coefficients to zero.

\setlength{\fboxsep}{0pt}

\renewcommand{\addimgT}[1]{\framebox{\includegraphics[width=0.073\linewidth]{#1.png}}}

\renewcommand{\addimgBigsingle}[1]{\includegraphics[width=0.22\linewidth]{image/single_compare_2/#1.png}}

\renewcommand{\AlgNametwo}[2]{\rotatebox{90}{\phantom{#2} \tiny  #1}}

\renewcommand{\addinputs}[1]{\raisebox{3em}{\multirow{3}{*}{ \begin{tabular}{c@{\hskip 0.4em}c}
\framebox{\includegraphics[width=0.2\linewidth]{#1im_crop.png}}\\ \addlinespace[0.2em]   Input (blur rate 0.44)  \end{tabular} } }}
\renewcommand{\addmetro}[1]{\raisebox{3em}{\multirow{3}{*}{ \begin{tabular}{c@{\hskip 0.4em}c}
\framebox{\includegraphics[width=0.2\linewidth]{#1metro_pred_0.png}}\\ \addlinespace[0.2em]    METRO~\cite{metro}  \end{tabular} } }}
\renewcommand{\addimgcropsyn}[1]{\framebox{\includegraphics[trim={0.4cm 1cm 1.0cm 0.4cm},clip,width=0.073\linewidth,height=0.073\linewidth]{#1.png}}}
\renewcommand{\addimgcropmean}[1]{\framebox{\includegraphics[trim={0.2cm 1cm 0.9cm 0.4cm},clip,width=0.073\linewidth,height=0.073\linewidth]{#1.png}}}

\renewcommand{\makeRowSingle}[1]{  
\addinputs{#1} & \AlgNameMu{GT +\cite{metro}}{25pt} & \addimgT{#1gt_0} & \addimgT{#1gt_1} & \addimgT{#1gt_2} & \addimgT{#1gt_3} & \addimgT{#1gt_4}&  \addimgT{#1gt_5}&  \addimgT{#1gt_6} \\
                 &   & \addimgT{#1gt_a_0} & \addimgT{#1gt_a_1} & \addimgT{#1gt_a_2} & \addimgT{#1gt_a_3} & \addimgT{#1gt_a_4} & \addimgT{#1gt_a_5} &  \addimgT{#1gt_a_6}   \\ \addlinespace[0.2em]
                 & \AlgNameMu{HfB (ours)}{35pt}  & \addimgT{#1hfb_render_new_0} & \addimgT{#1hfb_render_new_1} & \addimgT{#1hfb_render_new_2} & \addimgT{#1hfb_render_new_3} & \addimgT{#1hfb_render_new_4}&  \addimgT{#1hfb_render_new_5}&  \addimgT{#1hfb_render_new_6} \\ \addlinespace[0.2em]
  \addmetro{#1}  &  \AlgNameMu{AfB~\cite{zhong2022animation}+\cite{metro}}{30pt}       & \addimgT{#1afb_s_0} & \addimgT{#1afb_s_1} & \addimgT{#1afb_s_2} & \addimgT{#1afb_s_3} & \addimgT{#1afb_s_4}&  \addimgT{#1afb_s_5}&  \addimgT{#1afb_s_6} \\               
  &      & \addimgT{#1afb_a_0} & \addimgT{#1afb_a_1} & \addimgT{#1afb_a_2} & \addimgT{#1afb_a_3} & \addimgT{#1afb_a_4} & \addimgT{#1afb_a_5} &  \addimgT{#1afb_a_6}  \\ \addlinespace[0.1em]                  
                 &  \AlgNameMu{Jin~\etal~\cite{Jin_2018_CVPR} + \cite{metro}}{35pt}        & \addimgT{#1jin_s_0} & \addimgT{#1jin_s_1} & \addimgT{#1jin_s_2} & \addimgT{#1jin_s_3} & \addimgT{#1jin_s_4}&  \addimgT{#1jin_s_5}&  \addimgT{#1jin_s_6} \\
                 &   & \addimgT{#1jin_a_0} & \addimgT{#1jin_a_1} & \addimgT{#1jin_a_2} & \addimgT{#1jin_a_3} & \addimgT{#1jin_a_4} & \addimgT{#1jin_a_5} &  \addimgT{#1jin_a_6}   }
\begin{figure*}
\centering
\small
\setlength{\tabcolsep}{0.05em} 
\renewcommand{\arraystretch}{0.2} 
\begin{tabular}{@{}c@{\hskip 0.8em}cccccccc@{}}
\makeRowSingle{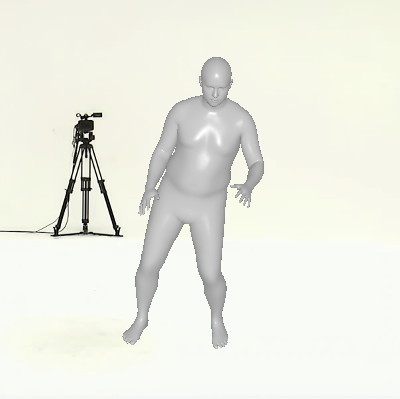} \\
  & & \multicolumn{7}{c}{Sub-frames}\\
\end{tabular}
\vspace{0.3em}
\caption{\textbf{Comparison on real data.}
We evaluate our method on the real B-AIST++~\cite{zhong2022animation} dataset. This example shows an average of 3 frames (see Table~\ref{tab:baist}). Our method produces more consistent and accurate sub-frame human poses compared to carefully selected baselines. }
\label{fig:synth_sfb}\end{figure*}


\section{Experiments}

Among the chosen baselines, we selected SfB \cite{sfb} and MfB~\cite{mfb}, designed for simple objects sub-frame deblurring and 3D reconstruction.
Then, we compare our method to general temporal super-resolution methods, \ie Jin~\etal~\cite{Jin_2018_CVPR} and Animation-from-Blur (AfB)~\cite{zhong2022animation} for single frame experiments and Blurry Video Frame Interpolation (BIN)~\cite{BIN} for multi-frame experiments.
To make them competitive, we also apply 3D human pose estimation METRO~\cite{metro} on top of their deblurred sub-frames (temporal super-resolution), except for SfB and MfB, where the output sub-frames are of low quality, and human pose estimation methods do not detect anything.

\boldparagraph{Blur rate.}

In general, motion blur is determined by many factors. 
However, the main factors are the camera exposure time and the speed of the object motion. 
Even with those two factors, it is still a challenging task to quantify the exact amount of motion blur. 
In order to measure the approximate blur level, we define blur rate as:
\begin{equation}
    \text{blur rate} = \frac{|\bigcup_{\tm=1}^{N} \rens(\mesh)|_1 }{|\rens(\meshi{1})|_1} -1 \enspace.
\end{equation}
Here, we compute the union of all projected sub-frame silhouettes and divide it by the first silhouette.
When the human stays still, the blur rate value is zero.
When the human moves over a distance larger than its size within one blurry frame, \ie there is no overlap between the rendered silhouettes at the first and last timestamps, the blur rate is larger than one.
We use this blur rate to classify the experiments.

\begin{table}[h]
\centering
\scriptsize
\setlength{\tabcolsep}{5.8pt} 
\newcommand{\MySkip}{\hskip 1.3em}  

\begin{tabular}{lcccccc}
  \toprule
         &  original &  2 avg. frames & 3 avg. frames \\    \midrule
  avg. blur rate       &    0.27   &  0.36     & 0.45 \\
  Method &   PA-MPJPE(mm)$\downarrow$      &      PA-MPJPE(mm)$\downarrow $         &    PA-MPJPE(mm)$\downarrow$           \\ 
  \midrule
   HfB  (ours)      & 69.1 & 77.3 & \textbf{81.4} \\     
   AfB~\cite{zhong2022animation}        & \textbf{52.3} & \textbf{63.3} & 87.3 \\
   Jin~\etal \cite{Jin_2018_CVPR} &   55.3   & 81.6 & 96.1 \\
  \bottomrule
\end{tabular}
\vspace{0.2em}
\caption{\textbf{Results on B-AIST++~\cite{zhong2022animation} dataset.} We average 2 and 3 original blurry frames to increase the blur amount. We apply METRO~\cite{metro} on top of the output of two baselines~\cite{Jin_2019_CVPR,zhong2022animation}. }
\label{tab:baist}

\end{table}

\subsection{Synthetic datasets}
%
We generated two datasets: BC-CAPE (Blur-Clothed CAPE~\cite{CAPE:CVPR:20}) and BT-AMASS (Blur-Textured AMASS~\cite{AMASS:2019}). 
The BT-AMASS is sampled on real-world human poses $\pose$, rotations $\rot$, and translations $\tran$ from the ACCAD~\cite{ACCAD} and CMU~\cite{cmuWEB} dataset of the AMASS~\cite{AMASS:2019} database with 120~fps. 
The UV textures $\texture$ are sampled from the SURREAL~\cite{surreal} dataset. 
Finally, the background images are randomly selected from a set of random images from the BG-20K~\cite{bg20k} database, capturing both indoors and outdoors scenes. 
We take random motion captures with length of 5 to 60 frames. 
This covers blur rates in the range between 0.05 and 1.1.
We utilize the SMPL-X plugin~\cite{SMPL-X:2019} in Blender to generate dataset images.

The BC-CAPE is based on the CAPE dataset~\cite{CAPE:CVPR:20}, which contains SMPL human models with poses for each frame with 60~fps. 
For BC-CAPE, we interpolate human poses to render higher speed footage. 
The camera position is randomly selected facing the human model. 
Then, we render sub-frame silhouettes $\rens(\mesh)$ and appearances $\renf(\mesh)$. 
In the end, we average these rendered silhouettes and appearances to acquire blurry images according to the image formation model~\eqref{eq:formation}. 
The jittering effect is eliminated by up-sampling and interpolation at a high frame rate of 600~fps. 
In total, we generated 1861 blurry images for a single frame experiment and 305 short videos with two video frames to evaluate our multi-frame setting.

\boldparagraph{Evaluation metrics.}
We evaluate HfB on the joint position error in millimeters: Mean Per Joint Position Error (MPJPE) and  Procrustes Analysis MPJPE (PA-MPJPE) as in~\cite{song2020lgd}. For MPJPE, we initially align the coordinate axis orientation of the predicted motion sequence with the ground truth. For comparison to SfB and MfB, we also measure the intersection-over-union (IoU) between the generated silhouette and the ground truth one.

\subsection{Results on BT-AMASS}
First, we compare our method on the generated BT-AMASS dataset to the following baselines: SfB~\cite{sfb}, MfB~\cite{mfb}, and static interpolated METRO~\cite{metro}
(1124 single frames and 305 two-frames). 
The single-frame results with 1124 images are shown in Table~\ref{tab:single_results},
whereas multi-frame results in Table~\ref{tab:multi_results}.

Our method outperforms all baselines by a wide margin, especially for higher blur rate intervals.
As excepted, the performance steadily decreases with the increased blur rate.
Additionally, we evaluate the influence of the Adversarial Motion Prior (AMP), which is only used for single-frame experiment.
This prior improves results only for higher blur raters, whereas for lower blur raters, where there is less ambiguity, it is harmful.

\renewcommand{\AlgNameMu}[2]{\multirow{1}{*}[#2]{\rotatebox{90}{\scriptsize #1 \phantom{Iy}}}}

\newcommand{\addimgb}[1]{\framebox{\includegraphics[width=0.12\linewidth]{#1}}}
\newcommand{\addimg}[1]{{\includegraphics[width=0.08\linewidth]{#1}}}


\newcommand{\makeRowSinHfB}[1]{\addimg{#10} & \addimg{#11} & \addimg{#12} & \addimg{#13} &  \addimg{#14}}
\newcommand{\makeRowSin}[1]{\addimg{#10} & \addimg{#11} & \addimg{#12} & \addimg{#13} &  \addimg{#14} & \addimg{#15} & \addimg{#16}  & \addimg{#17} & \addimg{#18} & \addimg{#19}}

\newcommand{\makeRowSinCrop}[1]{\addimgcrop{#10} & \addimgcrop{#11} & \addimgcrop{#12} & \addimgcrop{#13} &  \addimgcrop{#14} & \addimgcrop{#15} & \addimgcrop{#16}  & \addimgcrop{#17} & \addimgcrop{#18} & \addimgcrop{#19}}

\newcommand{\makeMRow}[1]{%
\AlgNameMu{Inputs}{30pt}  &\multicolumn{3}{c}{\addimgb{#1im_crop_0}} &  \multicolumn{3}{c}{\addimgb{#1im_crop_1}}    &  \multicolumn{3}{c}{\addimgb{#1im_crop_2}}  & \multicolumn{3}{c}{\addimgb{#1im_crop_3}}  \vspace{0.5em}\\  

\AlgNameMu{GT}{30pt}  & \addimg{#1gt_1} &\addimg{#1gt_2} &\addimg{#1gt_3} &   \addimg{#1gt_4} &\addimg{#1gt_5}&\addimg{#1gt_6}& \addimg{#1gt_7}&\addimg{#1gt_8}&\addimg{#1gt_9}&\addimg{#1gt_10}&\addimg{#1gt_11}&\addimg{#1gt_12}\\
\AlgNameMu{HfB}{30pt} & \addimg{#1body_0_1} &\addimg{#1body_0_2} &\addimg{#1body_0_3}& \addimg{#1body_1_1}&\addimg{#1body_1_2}&\addimg{#1body_1_3}& \addimg{#1body_2_1}&\addimg{#1body_2_2}&\addimg{#1body_2_3}&\addimg{#1body_3_1}&\addimg{#1body_3_2}&\addimg{#1body_3_3}\\
 & \addimg{#1bin_sharp_1}&\addimg{#1bin_sharp_2}&\addimg{#1bin_sharp_3}& \addimg{#1bin_sharp_4}&\addimg{#1bin_sharp_5}& \addimg{#1bin_sharp_6} & \addimg{#1bin_sharp_7} & \addimg{#1bin_sharp_8}& \addimg{#1bin_sharp_9}&\addimg{#1bin_sharp_10} & \addimg{#1bin_sharp_11}& \addimg{#1bin_sharp_12}\\
\AlgNameMu{BIN\cite{BIN} + METRO\cite{metro}}{80pt} & \addimg{#1bin_anno_1}& \addimg{#1bin_anno_2} &\addimg{#1bin_anno_3}& \addimg{#1bin_anno_4} &\addimg{#1bin_anno_5} &\addimg{#1bin_anno_6}& \addimg{#1bin_anno_7} & \addimg{#1bin_anno_8} & \addimg{#1bin_anno_9} &\addimg{#1bin_anno_10} & \addimg{#1bin_anno_11} & \addimg{#1bin_anno_12} \\
}

\begin{figure*}
\centering
\small
\setlength{\tabcolsep}{0.1em} 
\renewcommand{\arraystretch}{0.1} 
\begin{tabular}{@{}c@{\hskip 0.1em}ccc|@{\hskip 0.1em}ccc|@{\hskip 0.1em}ccc|@{\hskip 0.1em}ccc@{}}
&\multicolumn{3}{c}{ Frame \#1 } &  \multicolumn{3}{c}{ Frame \#2 }    &  \multicolumn{3}{c}{ Frame \#3}  & \multicolumn{3}{c}{ Frame \#4}   \\ \addlinespace[0.2em]
\makeMRow{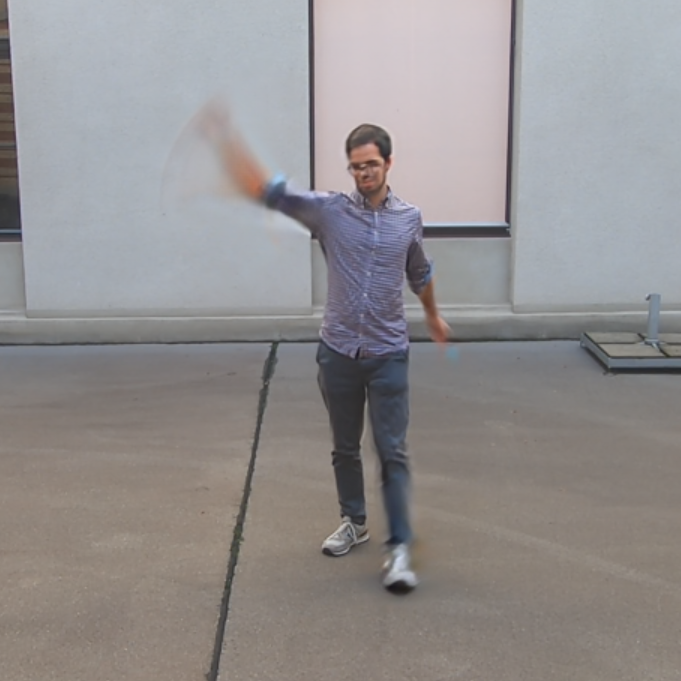}
\addlinespace[0.2em]
 & \MakeArrowVar{3} & \MakeArrowVar{3} & \MakeArrowVar{3} &\MakeArrowVar{3}\\\addlinespace[0.3em]
   & \multicolumn{3}{c}{Subframes of \#1 } & \multicolumn{3}{c}{Subframes of \#2 }  & \multicolumn{3}{c}{Subframes of \#3 } & \multicolumn{3}{c}{Subframes of \#4 }\\
\end{tabular}
\vspace{-0.1em}
\caption{\textbf{Multi-frame evaluation.} We compared to BIN~\cite{BIN}, with METRO~\cite{metro} human poses on top of their sub-frames. The visual results show that BIN fails, however METRO is still robust to some amount of blur and detects human poses, which are not consistent over time. The proposed method generates motion which is more consistent with the ground truth.  }
\label{fig:multi}
\end{figure*}

\begin{table*}[!htbp]
\centering
\scriptsize
\setlength{\tabcolsep}{1.0pt} 
\renewcommand{\arraystretch}{1.1} 
\newcommand{\NewSkip}{\hskip 1.5em}
\begin{tabular}{c@{\NewSkip}cc@{\NewSkip}cc@{\NewSkip}cc@{\NewSkip}cc@{\NewSkip}cc@{\NewSkip}cc}
\toprule
blur rate                                                       & \multicolumn{2}{c}{\textless{}0.2}       & \multicolumn{2}{c}{{[}0.2,0.3{]}}       & \multicolumn{2}{c}{{[}0.3,0.4{]}}       & \multicolumn{2}{c}{{[}0.4,0.5{]}}       & \multicolumn{2}{c}{{[}0.5,0.6{]}}       & \multicolumn{2}{c}{{[}0.6,0.8{]}}        \\ \cmidrule(lr){2-3} \cmidrule(lr){4-5} \cmidrule(lr){6-7} \cmidrule(lr){8-9} \cmidrule(lr){10-11} \cmidrule(lr){12-13}
Method                                                          & \tiny PA-MPJPE $\downarrow$ & \tiny MPJPE$\downarrow$ & \tiny PA-MPJPE$\downarrow$ & \tiny MPJPE$\downarrow$ & \tiny PA-MPJPE$\downarrow$ & \tiny MPJPE$\downarrow$ & \tiny PA-MPJPE$\downarrow$ & \tiny MPJPE$\downarrow$ & \tiny PA-MPJPE$\downarrow$ & \tiny MPJPE$\downarrow$ & \tiny PA-MPJPE$\downarrow$ & \tiny MPJPE$\downarrow$ \\ \midrule
HfB (ours)                                                  & 75.2 & 81.2 & 76.4 & 83.0 & 84.6 & 94.2 & \textbf{89.7} & \textbf{100.5} & \textbf{96.2} & \textbf{110.6} & \textbf{98.5} & \textbf{114.6} \\
AfB~\cite{zhong2022animation} + METRO~\cite{metro}  & 82.4 & 89.1 & 84.1 & 89.2 & 90.6 & 100.8 & 105.3 & 119.9 & 107.5 & 133.3 & 112.8 & 135.2 \\
Jin~\etal~\cite{Jin_2019_CVPR} + METRO~\cite{metro} & \textbf{74.4} & \textbf{77.2} & \textbf{76.2} & \textbf{82.0} & \textbf{82.6} & \textbf{90.8} & 100.1 & 112.8 & 99.5 & 119.2 & 105.6 & 124.6 \\ 
Jin~\etal~\cite{Jin_2019_CVPR}+PyMaF~\cite{pymaf2021} &79.9 & 83.6 & 80.3 & 85.5 & 93.4 & 106.1 & 115.6 & 123.3 & 119.5 & 137.9 & 125.2 & 142.5 \\ 
AfB~\cite{zhong2022animation}+PyMaF~\cite{pymaf2021} & 83.5 & 87.0 & 86.7 & 91.42 & 93.7 & 109.8 & 113.1 & 124.6 & 117.1 & 141.8 & 127.4 & 147.4  \\
\bottomrule

\end{tabular}
\vspace{0.3em}
\caption{\textbf{ Results on Blurred-Clothed CAPE dataset. } Our method outpeforms competitive baselines on larger blur rates.}
  \label{tab:bc_cape_results}
\end{table*}

\begin{table*}[!htbp]
\centering
\scriptsize
\setlength{\tabcolsep}{5.0pt} 
\renewcommand{\arraystretch}{1.1} 
\begin{tabular}{c@{\hskip 1.5em}cc@{\hskip 1.5em}cc@{\hskip 1.5em}cc@{\hskip 1.5em}cc@{\hskip 1.5em}cc@{\hskip 1.5em}cc}
\toprule
blur rate          & \multicolumn{2}{c}{\textless{}0.2}       & \multicolumn{2}{c}{{[}0.2,0.3{]}}       & \multicolumn{2}{c}{{[}0.3,0.4{]}}       & \multicolumn{2}{c}{{[}0.4,0.5{]}}       & \multicolumn{2}{c}{{[}0.5,0.7{]}}       & \multicolumn{2}{c}{{[}0.7,0.9{]}}        \\ \cmidrule(lr){2-3} \cmidrule(lr){4-5} \cmidrule(lr){6-7} \cmidrule(lr){8-9} \cmidrule(lr){10-11} \cmidrule(lr){12-13}
Method & \tiny PA-MPJPE $\downarrow$ & \tiny MPJPE$\downarrow$ & \tiny PA-MPJPE$\downarrow$ & \tiny MPJPE$\downarrow$ & \tiny PA-MPJPE$\downarrow$ & \tiny MPJPE$\downarrow$ & \tiny PA-MPJPE$\downarrow$ & \tiny MPJPE$\downarrow$ & \tiny PA-MPJPE$\downarrow$ & \tiny MPJPE$\downarrow$ & \tiny PA-MPJPE$\downarrow$ & \tiny MPJPE$\downarrow$ \\ \midrule
Hfb (ours)  & 83.7  & 85.3 &  \textbf{86.1} & \textbf{91.3} & \textbf{90.5}   & \textbf{96.4}  & \textbf{92.6}  &  \textbf{99.3} &  \textbf{104.2}  & \textbf{114.4} &  \textbf{110.5}  & \textbf{117.4} \\
BIN~\cite{BIN} + HybrIK~\cite{li2021hybrik}      & \textbf{76.6}  & \textbf{78.8} &  86.7 & 96.5 & 93.5   & 106.9 & 107.0 & 122.9 &  116.3  & 134.5 &  120.2  & 150.0 \\
AfB~\cite{zhong2022animation} + HybrIK~\cite{li2021hybrik} & 82.4  & 84.6 &  96.8 & 105.3 &  100.8 & 111.1 &  103.6  & 120.1 &  119.1 & 137.9 &  120.9 & 156.2           \\     
BIN~\cite{BIN} + METRO~\cite{metro} & 84.0  & 85.9 &  88.5 & 97.5 &  94.0   & 106.8 &  100.1  & 118.4 &  112.8 & 133.7 &  117.4 & 146.4 \\
AfB~\cite{zhong2022animation} + METRO~\cite{metro} & 87.7 & 90.2 &  90.8 & 98.1 & 97.7 & 108.9 & 109.0 & 126.2 &113.0 & 133.3 & 121.8 & 151.2 \\ 
PyMaF~\cite{pymaf2021} & 97.7 & 119.5 & 118.7 & 152.5 & 128.8 & 172.0 & 138.3 & 202.0 & 152.7 & 231.4 & 157.9 & 252.9   \\
BIN~\cite{BIN}+PyMaF~\cite{pymaf2021}  & 78.29 & 98.04 & 95.05 & 126.2 & 105.2 & 151.2 & 110.7 & 147.6 & 121.7 & 168.4 & 124.0 & 187.7 \\
\bottomrule
\end{tabular}
\vspace{0.3em}
\caption{\textbf{ Results on Blurred-Clothed CAPE dataset with 4 consecutive frames.} We also combine both METRO~\cite{metro} and HybrIK~\cite{li2021hybrik} with two baselines (BIN~\cite{BIN} and AfB~\cite{zhong2022animation}) to show the impact of different human pose estimation methods. We also show results of BIN~\cite{BIN} with PyMaF~\cite{pymaf2021}, which include the interpolation of the joint positions by directly applying on blur frames. With multiple input frames, HfB outpeforms other baselines for almost all blur rates (0.2 and higher). }
  \label{tab:bc_cape_multi_results}
\vspace{-0.5em}
\end{table*}

\subsection{Results on BC-CAPE}

Next, we evaluate the proposed method on the generated BC-CAPE dataset, which contains 609 single frames and 205 short sequences with 4-frames.
In this case, we compare to three temporal super-resolution methods: AfB~\cite{zhong2022animation}, Jin~\etal~\cite{Jin_2018_CVPR}, and BIN~\cite{BIN}.
For fair comparison, we augmented their sub-frame output with human pose estimation metods, either METRO~\cite{metro} or HybrIK~\cite{li2021hybrik}. 
As shown in Tables \ref{tab:bc_cape_multi_results} and \ref{tab:bc_cape_results}, the proposed Human from Blur (HfB) method outperforms these baselines by a large margin, especially on larger blur rates.
The performance gain is even higher for the multi-frame experiment (Table~\ref{tab:bc_cape_multi_results}).

\subsection{Real dataset B-AIST++}
Finally, we evaluate on the real-world dataset with various human motion and garments: B-AIST++~\cite{zhong2022animation}.  
They use frame interpolation to generate high-speed frames from original dancing dataset AIST++~\cite{li2021learn}.
B-AIST++ provides significantly blurred images with human motion.
We generate ground-truth sub-frame human pose by running METRO~\cite{metro} on top of the ground-truth sub-frames.
Table~\ref{tab:baist} shows that our method outperforms other baselines when 3 consecutive frames are averaged, which translates to blur rate 0.45. 
Note that AfB~\cite{zhong2022animation} is trained on this dataset, whereas our method is purely optimization based.

\subsection{Captured data}

We captured 21 real-world sequences with significant amounts of motion blur, including four male and one female subjects.
The used cameras are the IDS camera and a GoPro 7, which were deliberately set at a low frame rate of 30 fps with exposure time of 30 ms to 50 ms.
The recorded humans were asked to move fast.
Background images are captured as well.
As shown in Fig.~\ref{fig:real} and Fig.~\ref{fig:teaser}, the final reconstructions are plausible.
When compared to SfB~\cite{sfb}, our method achieves significantly better results.

\section{Conclusion}
%
We proposed the first method to reconstruct sub-frame human motion and textured shape from substantially blurred images.
The key idea is to approach the problem from a generative viewpoint and describe a fully differentiable forward process to generate blurry images from a given 3D human motion model. 
The core of our method is an image reconstruction loss that allows to solve the inverse problem with standard gradient descent methods.
Experiments showed that the proposed method achieves the best results on both synthetic and real blurry data.

\vspace{1em}
{\textbf{Acknowledgements.} This research was supported by a Google Focused Research Award, Innosuisse grant No.~34475.1 IP-ICT, and a research grant by FIFA.}

{\small
\bibliographystyle{ieee_fullname}
\bibliography{egbib}
}
\end{document}